%% file: main.tex
\documentclass[11pt]{article}

\usepackage[margin=1in]{geometry}
\usepackage[utf8]{inputenc}
\usepackage[T1]{fontenc}
\usepackage{textcomp}
\usepackage{amsmath}
\usepackage{amsfonts}
\usepackage{amssymb}
\usepackage{nicefrac}
\usepackage{microtype}
\usepackage{xcolor}
\usepackage{graphicx}
\usepackage{booktabs}
\usepackage{multirow}
\usepackage{colortbl}
\usepackage{tabularx}
\usepackage{siunitx}
\usepackage{algorithm}
\usepackage{algpseudocode}
\usepackage{subcaption}
\usepackage{enumitem}
\usepackage{placeins}
\usepackage{wrapfig}
\usepackage{parskip}
\usepackage{etoolbox}
\usepackage[round]{natbib}
\usepackage{url}
\usepackage[colorlinks=true,linkcolor=blue,citecolor=blue,urlcolor=blue]{hyperref}

\newcommand{\arxivmode}{}

\newcommand{\R}{\mathbb{R}}
\newcommand{\loss}{\mathcal{L}}

\newcommand{\frobnorm}[1]{\lVert #1 \rVert_{\mathrm{F}}}
\DeclareMathOperator*{\argmax}{arg\,max}

\newcommand{\protok}[1]{ProTok\textsubscript{\texttt{#1}}}

\AtBeginEnvironment{table}{\let\scriptsize\footnotesize}
\AtBeginEnvironment{table*}{\let\scriptsize\footnotesize}

\renewenvironment{abstract}{\begin{center}\bfseries\abstractname\end{center}\vspace{-0.5em}}{\par}

\title{ProTok: Unified Protein Tokenization via Modular Multi-Codebook Quantization}
\author{Romain Hardy \qquad Xiaoman Zhang \qquad Pranav Rajpurkar \\[12pt] \textit{Department of Biomedical Informatics, Harvard Medical School}}
\date{}

\begin{document}

\maketitle

\begin{abstract}
\input{content/00_abstract}
\end{abstract}

\input{content/01_intro}
\input{content/02_related}
\input{content/03_method}
\input{content/04_experiments}
\input{content/05_results}
\input{content/06_future}

\bibliographystyle{plainnat}
\bibliography{references}

\clearpage
\appendix
\input{content/appendix}

\end{document}

%% file: content/00_abstract.tex
Protein sequence, structure, and functional annotations each capture complementary aspects of protein biology, yet existing tokenizers encode each modality independently, limiting the expressiveness of discrete protein representations.
We present \ifdefined\arxivmode ProTok\else\textbf{ProTok}\fi, a unified protein tokenizer that maps all three modalities into a modular discrete vocabulary via product quantization with per-modality codebook groups and modality dropout.
A single ProTok checkpoint supports flexible inference from any modality subset without retraining.
On seven StructTokenBench benchmarks spanning 14 tasks and 28 test splits, ProTok achieves an average rank of 1.4 across all structure-only tokenizers, placing first on 21 splits and among the top two on 27 of the 28.
These gains come without sacrificing reconstruction: ProTok's tokens reach 0.939 lDDT on CATH~4.2, showing that multi-codebook quantization and multi-modal training yield tokens that are both reconstructable and informative.
ProTok's structure tokens further lead all baseline structure tokenizers on zero-shot functional retrieval across CATH, Enzyme Commission, and Gene Ontology label hierarchies.
The vocabulary also supports generation: a compact language model trained from scratch over ProTok tokens jointly generates backbone and sequence, with 52\% of sampled backbones designable versus 21\% for a parameter-matched GeoBPE baseline.
ProTok shows that structure, sequence, and function can share one discrete vocabulary, motivating protein language models that reason over a unified token stream.

%% file: content/01_intro.tex
\section{Introduction}
\label{sec:introduction}

Proteins are characterized by amino acid sequences that fold into three-dimensional structures to carry out diverse biological functions. Understanding and modeling proteins computationally requires reasoning across all three levels: the structure determines molecular interactions, the sequence encodes evolutionary constraints, and functional annotations describe biological roles, binding properties, and phenotypic effects. Recent protein language models~\citep{elnaggar2021prottrans,lin2023evolutionary,hayes2024simulating,wang2024dplm2,ye2024proteinbench} have made substantial progress by learning from these information sources, but the tokenizers they rely on still encode each modality independently.

\begin{figure*}[t]
\centering
\includegraphics[width=\textwidth]{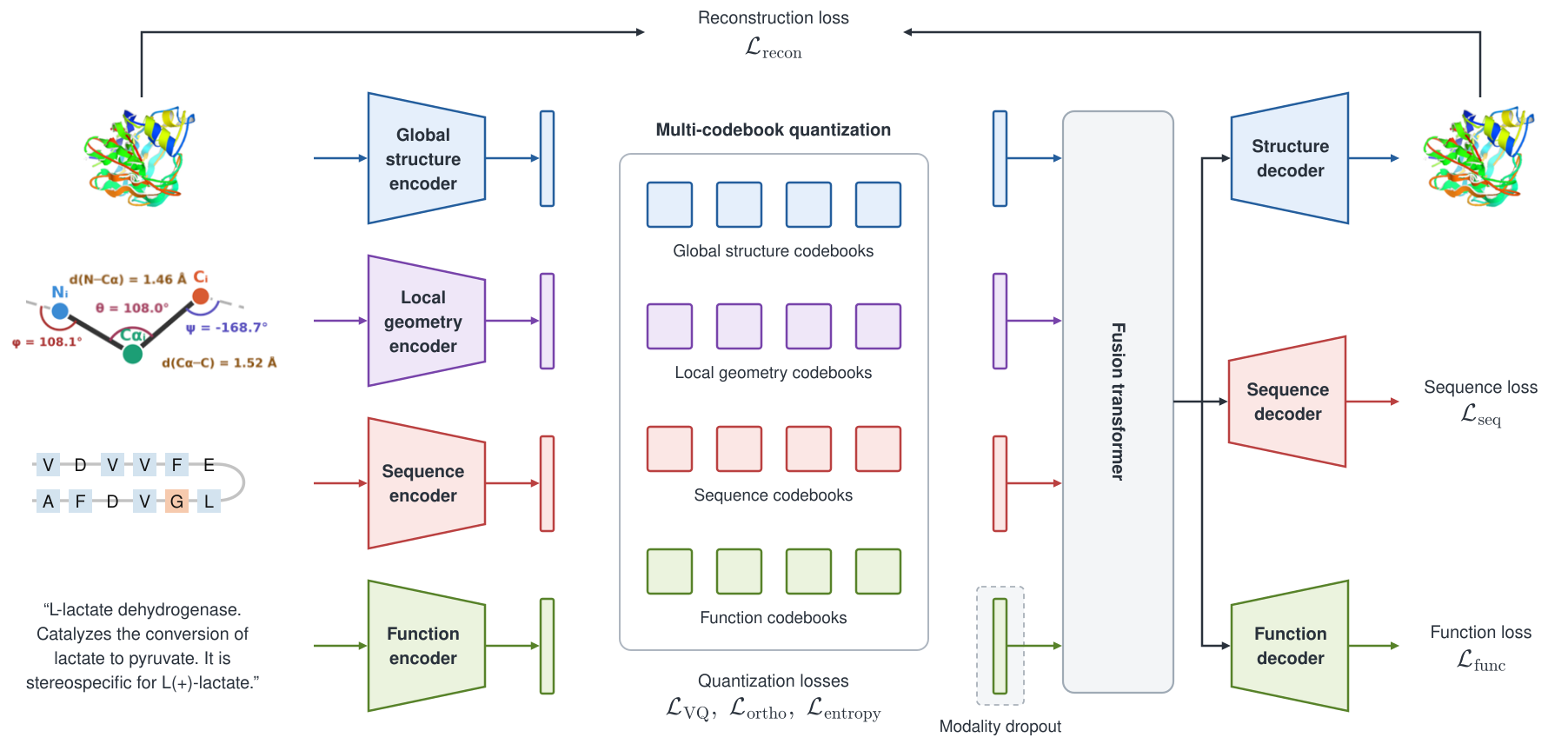}
\caption{\textbf{ProTok architecture.} Four frozen encoders process global structure, local geometry, sequence, and function inputs. Each encoder output is quantized through a dedicated group of codebooks. During training, modality dropout randomly masks groups and replaces them with learnable mask tokens (illustrated on the function group), forcing the structure decoder and the auxiliary sequence and function decoders to reconstruct from any modality subset. A single-layer fusion module attends across active groups before decoding. At inference, the same checkpoint operates in different modes by selecting which groups are active.}
\label{fig:architecture}
\end{figure*}

We identify two limitations of this independent encoding. First, \textbf{existing protein tokenizers are typically designed for a single modality}: structure tokenizers~\citep{van2024fast,guo2024structtokenbench,chen2024prosst} discard evolutionary signal, while sequence tokenizers~\citep{lin2023evolutionary} capture limited geometric information. The rare attempts at joint sequence-structure tokenization~\citep{gao2025foldtoken,su2023saprot} omit functional supervision; no existing tokenizer jointly quantizes structure, sequence, and function into a modular vocabulary that supports flexible inference from any modality subset. Second, \textbf{single-codebook quantizers~\citep{van2017neural} widely used in structure tokenizers suffer from codebook collapse at scale}, where only a fraction of code vectors receive meaningful gradient updates, shrinking the effective vocabulary and limiting representational expressiveness. Multi-codebook product quantization~\citep{jegou2010product,ma2025unitok} alleviates the capacity problem, but existing designs remain modality-specific.

We propose \textbf{ProTok}, a unified protein tokenizer that addresses both limitations, mapping sequence, structure, and function into a single cross-modal vocabulary that serves as a substrate for protein language modeling and structure-sequence co-generation. We make three contributions:
\begin{itemize}[topsep=2pt,itemsep=3pt,parsep=0pt]
    \item \textbf{A unified multimodal tokenizer.} ProTok extracts representations from four pretrained encoders~\citep{hayes2024simulating,guo2024structtokenbench,hsu2022learning,lin2023evolutionary,qwen3} capturing global structural context, local backbone geometry, evolutionary sequence features, and protein-level functional semantics. Each encoder output is quantized through a dedicated group of 4 codebooks, yielding a total of 16 parallel codes per residue. Orthogonality-regularized SimVQ projections keep each codebook full-rank, so the product code's capacity remains usable. The quantized outputs are fused via cross-group attention and decoded to reconstruct backbone geometry, recover amino acid identity, and align with frozen sequence and functional embeddings.
    \item \textbf{Any-subset inference from a single checkpoint.} During training, codebook groups are randomly masked and replaced with learnable mask tokens, forcing the shared decoders to recover their targets from any active modality subset. This enables a single trained checkpoint to operate in structure-only, sequence-only, or combined mode at inference without retraining.
    \item \textbf{One vocabulary, three capabilities.} Among the tokenizers we evaluate, ProTok is the strongest overall on StructTokenBench, and its frozen tokens lead every baseline on zero-shot functional retrieval; the same tokens reconstruct structure with high fidelity and form a generative substrate from which a language model co-generates designable protein backbones and sequences. To our knowledge, no prior tokenizer demonstrates this combination.
\end{itemize}

We train ProTok in two stages on 311K protein chains from the Protein Data Bank (PDB)~\citep{berman2000protein} and AlphaFold~2-predicted Swiss-Prot~\citep{jumper2021highly,uniprot2023}, and evaluate on StructTokenBench~\citep{guo2024structtokenbench} across seven benchmarks spanning functional site prediction, flexibility regression, remote homology classification, and conformational discrimination.
ProTok achieves the strongest overall profile among structure-only tokenizers, with a mean rank of 1.4 across 28 test splits, placing first on 21 splits and among the top two on 27 of the 28.
The same tokens are reconstructable and generative: they reconstruct backbone structure at 0.939 lDDT on CATH~4.2, and a language model over ProTok tokens jointly generates backbone and sequence, with 52\% of generated backbones designable by self-consistency versus 21\% for a parameter-matched GeoBPE baseline.
Two further probes confirm that these tokens carry cross-modal signal: ProTok's structure tokens lead every baseline on zero-shot functional retrieval across CATH, Enzyme Commission, and Gene Ontology hierarchies, and activating additional codebook groups at inference monotonically improves functional site prediction.

%% file: content/02_related.tex
\section{Related Work}
\label{sec:related}

\paragraph{Protein structure tokenizers.}
Discrete tokenization of protein structure has largely converged around per-residue VQ-VAE codebooks trained to reconstruct backbone geometry. ESM3~\citep{hayes2024simulating} demonstrates that this paradigm can reach near-perfect reconstruction at scale, yet its tokens underperform on semantic downstream tasks. AminoAseed~\citep{guo2024structtokenbench} instead employs codebook reparameterization and Pareto-optimal sizing, recovering substantial downstream performance at the cost of weaker reconstruction. DPLM-2~\citep{wang2024dplm2} applies lookup-free quantization to GVP-Transformer~\citep{hsu2022learning,jing2021learning} encoder embeddings, growing the effective vocabulary without enumerating codebook entries. ProSST~\citep{chen2024prosst} takes a different route, replacing learned quantization with $k$-means clustering over a frozen GVP encoder. FoldSeek~\citep{van2024fast} goes further still, mapping residue-neighbor geometry to a fixed 20-state 3Di alphabet originally designed for fast structure search. Beyond these alternatives to learned per-residue VQ, several recent variants extend tokenization along different design axes: SE(3)-equivariant encoders (GCP-VQVAE~\citep{pourmirzaei2025gcpvqvae,morehead2024gcpnet}), motif-level granularity (GeoBPE~\citep{sun2026geobpe,sennrich2016neural}), joint sequence-structure quantization (FoldToken~\citep{gao2025foldtoken}), all-atom tokenization across biomolecule types (Bio2Token~\citep{liu2024bio2token}), and, in recent work, flow-matching reconstruction objectives (ProFiT~\citep{zhang2026profit}). While some of these systems incorporate multiple modalities downstream by feeding their structure tokens into a multi-modal language model, the tokenization pipelines themselves remain modality-specific; no existing tokenizer provides a unified encoding of structure, sequence, and function within a single vocabulary. The StructTokenBench framework~\citep{guo2024structtokenbench} provides systematic evaluation across seven benchmarks and serves as our primary evaluation protocol.

\vspace{-1pt}
\paragraph{Multi-codebook and product quantization.}
Product quantization~\citep{jegou2010product} decomposes a high-dimensional vector into independently quantized subvectors, yielding an exponentially large effective vocabulary from compact per-codebook lookups. SimVQ~\citep{zhu2024scalingcodebooksizevqmodels} addresses codebook collapse, in which most entries receive no gradient updates, by reparameterizing each codebook as a frozen random embedding composed with a trainable linear projection that moves all entries coherently. In the vision domain, UniTok~\citep{ma2025unitok} shows that multi-codebook quantization can produce a single visual vocabulary that serves both image generation and understanding. ProTok adapts these ideas to proteins, adding an orthogonality penalty~\citep{shin2023translation}, originally applied to codebook embeddings, which we instead impose on the SimVQ projection matrices to prevent rank collapse, and a modular codebook design with dedicated groups for global structure, local geometry, sequence, and function.

\vspace{-1pt}
\paragraph{Multi-modal protein representations.}
Existing multi-modal protein systems combine structure and sequence in one of two ways. ESM3~\citep{hayes2024simulating} and DPLM-2~\citep{wang2024dplm2} maintain a parallel token stream for each modality (a dedicated structure tokenizer alongside the standard amino acid vocabulary) and rely on the language model to mix them via attention. SaProt~\citep{su2023saprot} and CHEAP~\citep{lu2025tokenized} instead collapse the two modalities into a single per-residue representation: SaProt forms a joint structure-aware alphabet by concatenating FoldSeek 3Di states with amino acid identity, while CHEAP replaces discrete quantization with a continuous bottleneck over both modalities. None of these methods' tokenizers are trained with cross-modal incentives, and none extend to functional supervision. Concurrent with our work, Yeti~\citep{giri2026yeti} pairs a compact lookup-free structure tokenizer with a masked-diffusion model trained jointly over its structure tokens and amino acid sequences to co-generate backbones and sequences; as in ESM3 and DPLM-2, its tokenizer is structure-only, and it does not model function. ProTok instead quantizes structure, sequence, and function through a single multi-codebook architecture in which modality dropout forces each group to encode information that supports reconstruction even when others are masked; a single trained checkpoint can therefore be applied to any subset of the modalities at inference, yielding a vocabulary whose tokens carry signal beyond their nominal modality.

%% file: content/03_method.tex
\section{ProTok}
\label{sec:protok}

ProTok partitions the quantization stage into modality-specific codebook groups, fuses their outputs through a cross-group attention module, and applies modality dropout during training to ensure robust reconstruction from any subset of available inputs.
The overall framework is illustrated in Figure~\ref{fig:architecture}.

\subsection{Encoders}
\label{sec:encoders}
ProTok extracts features from four pretrained encoders, with their backbones frozen throughout training, ensuring that each modality retains the representation quality of its source model.
Each encoder output is projected to a shared $d = \textrm{1024}$-dimensional space via a trainable two-layer MLP.

\paragraph{Global structure encoder.}
We use ESM3's two-layer geometric attention transformer~\citep{hayes2024simulating}, initialized from the AminoAseed checkpoint~\citep{guo2024structtokenbench}.
Each layer attends over each residue's $k = \textrm{16}$ nearest $\text{C}_\alpha$ neighbors, with attention scores modulated by the relative geometry of local SE(3) frames.

\paragraph{Local geometry encoder.}
A frozen ESM-IF1~\citep{hsu2022learning} GVP4 encoder~\citep{jing2021learning} captures local backbone geometry, complementing the global encoder's learned representations.

\paragraph{Sequence encoder.}
A frozen ESM-2 650M model~\citep{lin2023evolutionary} provides per-residue evolutionary features learned from protein sequence databases, capturing conservation and co-evolutionary patterns not directly accessible from a single 3D structure.

\paragraph{Function encoder.}
Text annotations from UniProt (protein names, function descriptions, GO terms, keywords) are encoded by a frozen Qwen3-Embedding 0.6B model~\citep{qwen3}, providing a single protein-level embedding that captures functional context beyond what can be inferred from structure or sequence alone.
This embedding is quantized once per protein, and the resulting codes are broadcast to all residue positions so that the function group produces per-residue codes with the same shape as the other modalities; the function loss operates on mean-pooled quantized representations to match the protein-level nature of the source signal.

\subsection{Quantization}
\label{sec:mcq}

\paragraph{Multi-codebook vector quantization.}
The quantization stage is partitioned into $G = \textrm{4}$ modality-specific codebook groups (global structure, local geometry, sequence, and function).
Within each group, we apply product quantization~\citep{jegou2010product}, splitting the projected encoder output $\mathbf{z}_i \in \R^d$ at each residue $i = \textrm{1}, \ldots, L$ (with $d = \textrm{1024}$) into $K = \textrm{4}$ equal chunks $\mathbf{z}_i^{(k)} \in \R^{d_\mathrm{sub}}$ of dimension $d_\mathrm{sub} = d/K = \textrm{256}$ and quantizing each chunk against a codebook $\mathbf{C}^{(k)} \in \R^{C \times d_\mathrm{sub}}$ with $C = \textrm{256}$ entries (Figure~\ref{fig:architecture}).
Following SimVQ~\citep{zhu2024scalingcodebooksizevqmodels}, we parameterize each codebook as $\mathbf{C}^{(k)} = \mathbf{W}^{(k)} \mathbf{C}_\textrm{0}^{(k)}$, where $\mathbf{C}_\textrm{0}^{(k)}$ is a frozen random matrix and $\mathbf{W}^{(k)} \in \R^{d_\mathrm{sub} \times d_\mathrm{sub}}$ is a trainable linear projection; all entries of a codebook then move coherently under each gradient step, discouraging dead codes.
To keep the projections from collapsing onto low-rank subspaces (which would shrink the effective vocabulary), we collect the $GK$ projections across all groups and regularize each toward orthogonality, adopting a penalty~\citep{shin2023translation} originally applied to codebook embeddings and now imposed on the SimVQ projection matrices:
\begin{equation}
  \loss_\mathrm{ortho} = \frac{1}{GK} \sum_{m=\textrm{1}}^{GK}
  \frobnorm{\mathbf{W}_m^\top \mathbf{W}_m - \mathbf{I}}^\textrm{2},
  \label{eq:ortho}
\end{equation}
weighted by $\lambda_\mathrm{ortho} = \textrm{0.01}$.

\paragraph{Token assignment.}
Within each group, chunk $\mathbf{z}_i^{(k)}$ is matched to the nearest entry in $\mathbf{C}^{(k)}$ by maximum cosine similarity:
\begin{equation}
  q^{(k)}(i) = \argmax_{c \in \{\textrm{1},\ldots,C\}}
  \langle \bar{\mathbf{z}}_i^{(k)},\, \bar{\mathbf{C}}^{(k)}_c \rangle,
  \quad k = \textrm{1}, \ldots, K,
  \label{eq:token_assignment}
\end{equation}
where $\bar{\cdot}$ denotes $\ell_2$ normalization.
The group's quantized representation at residue $i$ is the concatenation of the $K$ selected entries, with gradients propagated via the straight-through estimator~\citep{van2017neural}.
Across the $G$ groups, ProTok assigns $GK = \textrm{16}$ discrete codes per residue.
These codes are fused into a single per-residue vector (Section~\ref{sec:decoder}), so a length-$L$ protein forms a stream of $L$ positions, not $\textrm{16}L$ tokens.

\subsection{Fusion and Reconstruction}
\label{sec:decoder}

\paragraph{Fusion.}
Within each group, the $K$ quantized codebook outputs are concatenated into a single vector per residue and projected to $\R^{d}$ by a trainable linear map.
A single-layer attention module then attends across the $G$ group representations at each residue position, and the attended vectors are summed to produce $\mathbf{z}_\mathrm{fused} \in \R^{L \times d}$.

\paragraph{Decoders.}
An 8-layer bidirectional transformer following the ESM3 (Stage 1) and AminoAseed decoder architecture takes $\mathbf{z}_\mathrm{fused}$ and reconstructs four geometric targets (pairwise backbone atom distances, backbone direction vectors, binned $\text{C}_\beta$ distances, and binned pairwise direction cosines) together with amino acid identity (inverse folding).
Two auxiliary decoders supplement the structure decoder: a sequence head (two-layer MLP predicting ESM-2 hidden states from $\mathbf{z}_\mathrm{fused}$) and a function head (two-layer MLP predicting the Qwen3 function embedding from mean-pooled $\mathbf{z}_\mathrm{fused}$).

\paragraph{Inference modes.}
A single ProTok checkpoint supports multiple inference modes by selecting which codebook groups are active.
For instance, \textbf{\protok{str}} uses the global structure and local geometry groups (8 codebooks); \textbf{\protok{str-seq}} adds the sequence group (12 codebooks); and
\textbf{\protok{str-seq-fun}} combines all groups (16 codebooks).
No retraining is required to switch modes.
For each mode, discrete token IDs index into the corresponding codebook embeddings, which are concatenated across the active codebooks to form feature vectors.

\subsection{Training}
\label{sec:training}

\paragraph{Objective.}
The training objective combines four loss components with two codebook regularizers.
The reconstruction loss $\loss_\mathrm{recon}$ supervises the five structure decoder outputs from Section~\ref{sec:decoder} with mean squared error on the two backbone geometry targets and cross-entropy on the two binned targets and the inverse folding prediction.
The sequence loss $\loss_\mathrm{seq}$ and function loss $\loss_\mathrm{func}$ align the quantized representation against the frozen ESM-2 hidden states and the mean-pooled Qwen3 protein embedding, respectively.
The VQ loss $\loss_\mathrm{VQ}$ is the standard combination of commitment and codebook alignment terms from VQ-VAE.
Finally, $\loss_\mathrm{ortho}$ (Equation~\ref{eq:ortho}) penalizes rank collapse in the SimVQ projections, and $\loss_\mathrm{entropy}$ encourages uniform code usage.
The full objective is:
\ifdefined\arxivmode
\begin{equation}
  \loss = \loss_\mathrm{recon} + \loss_\mathrm{seq} + \loss_\mathrm{func}
  + \loss_\mathrm{VQ} + \lambda_\mathrm{ortho}\,\loss_\mathrm{ortho} + \lambda_\mathrm{entropy}\,\loss_\mathrm{entropy}.
  \label{eq:total_loss}
\end{equation}
\else
\begin{equation}
\begin{split}
  \loss ={}& \loss_\mathrm{recon}
  + \loss_\mathrm{seq}
  + \loss_\mathrm{func} \\
  &+ \loss_\mathrm{VQ}
  + \lambda_\mathrm{ortho}\,\loss_\mathrm{ortho}
  + \lambda_\mathrm{entropy}\,\loss_\mathrm{entropy}.
\end{split}
\label{eq:total_loss}
\end{equation}
\fi

Appendix~\ref{app:training_details} reports per-target loss forms, binning schemes, and coefficient values.

\paragraph{Masking.}
ProTok uses two complementary masking mechanisms during training.
Each group $g$ has a single learnable mask token $\mathbf{m}_g \in \R^d$ (initialized to zero, so masked groups contribute nothing at the start of training) that serves both mechanisms below.
\emph{Modality dropout} operates at the sample level: for each protein, we uniformly sample one of the $\textrm{2}^G - \textrm{1} = \textrm{15}$ non-empty combinations of codebook groups to retain, replacing each dropped group's contribution with its mask token $\mathbf{m}_g$.
This forces the decoders to reconstruct from any modality combination and enables multi-mode inference from a single checkpoint.
\emph{Per-residue masking} operates at the residue level: 15\% of residue positions are masked uniformly at random after quantization, and the quantized representations at those positions are replaced by the same $\mathbf{m}_g$.
A single residue mask is shared across the global structure, local geometry, and sequence groups to prevent positional information leakage; per-residue masking does not apply to the function group, since the function encoder's protein-level embedding is identical across residues by construction.
Modality dropout creates diverse cross-modal training scenarios while per-residue masking forces the model to learn from spatial context rather than relying on per-position identity.

\paragraph{Training stages.}
We train ProTok in two stages, separating tokenizer learning from reconstruction-decoder learning as in ESM3~\citep{hayes2024simulating}.
\textbf{Stage~1} jointly optimizes the quantizer, fusion module, an 8-layer structure decoder, and the auxiliary sequence and function heads on 311K protein chains drawn from PDB~\citep{berman2000protein} (48K experimental chains) and Swiss-Prot~\citep{jumper2021highly,uniprot2023} (263K AlphaFold~2-predicted chains); we remove exact sequence matches and 50\%-identity sequence clusters overlapping any evaluation benchmark before training.
\textbf{Stage~2} freezes everything learned in Stage~1 (encoders, quantizers, codebook projections, and the Stage~1 sequence and function heads), replaces the 8-layer structure decoder with a larger 30-layer decoder initialized from the pretrained ESM3 structure decoder, and adds two new auxiliary heads that regress the pLDDT and predicted aligned error (PAE) signals of AlphaFold~2.
Both stages train for 25 epochs; Appendix~\ref{app:training_details} lists the optimizer and remaining hyperparameters used in our training recipe.

%% file: content/04_experiments.tex
\section{Experiments}
\label{sec:experiments}

\paragraph{Evaluating structural token quality.}
We assess ProTok's structure tokens from two complementary angles. 
Extrinsically, we evaluate on StructTokenBench~\citep{guo2024structtokenbench}, which groups seven benchmarks into 14 prediction tasks spanning functional site prediction, flexibility regression, remote homology classification, and conformational discrimination, and reports results on 28 held-out splits, including fold- and superfamily-level holdouts that test structural generalization. 
For each task, we train a lightweight classifier on frozen tokens and report test performance at the best validation checkpoint (seed variance in Appendix~\ref{app:seeds}); conformational tasks are evaluated zero-shot.
In the \protok{str} mode, only the structure groups are read out, so its tokens are directly comparable with those of every structure-only tokenizer; the sequence and function groups shape them only through joint training.
Intrinsically, we report codebook utilization rate (fraction of codes assigned to any residue), normalized perplexity (evenness of the code frequency distribution), and native reconstruction quality on the test split of CATH~4.2~\citep{sillitoe2021cath}.

\paragraph{Probing codebook information content.}
Three analyses probe what \protok{}'s codebooks encode: on CATH~4.2, standardized inverse-folding and structure-prediction probes that isolate token information content from each tokenizer's decoder capacity, and zero-shot functional retrieval over nine CATH, Enzyme Commission, and Gene Ontology label hierarchies; on StructTokenBench tasks, a nested-mode study (\protok{str}, \protok{str-seq}, \protok{str-seq-fun}; protocols in Appendix~\ref{app:eval_setup}).

\paragraph{Baselines.}
We compare against four discrete structure tokenizers spanning varied design choices: a single learned codebook (ESM3~\citep{hayes2024simulating}, AminoAseed~\citep{guo2024structtokenbench}), lookup-free quantization (DPLM-2~\citep{wang2024dplm2}), and $k$-means clustering of frozen geometric features (ProSST~\citep{chen2024prosst}).
We use each baseline's official released checkpoint under the same probe protocols and hyperparameter sweep as ProTok.

\paragraph{Assessing generation.}
To test whether ProTok's tokens form a usable generative substrate, we train a masked generative prior over \protok{str-seq} tokens on a CATH~4.2 corpus of chains up to 128 residues and evaluate sampled backbones by self-consistency, reporting designability, co-designability, diversity, and uniqueness (Appendix~\ref{app:generation}).
We benchmark generation against GeoBPE~\citep{sun2026geobpe}, which reports designability at a parameter budget matched to our prior; we exclude larger generative models such as ESM3 and DPLM-2, whose scale precludes a size-controlled comparison.

%% file: content/05_results.tex
\definecolor{rowshade}{gray}{0.85}
\definecolor{taskshade}{gray}{0.92}
\definecolor{protokhi}{HTML}{FFFBE6}

\ifdefined\arxivmode\begin{table*}[p]\else\begin{table*}[t]\fi
  \centering
  \caption{\textbf{Downstream evaluation on StructTokenBench.} We evaluate \protok{str} against discrete structure tokenizers on functional site prediction (AUROC\,\%), flexibility and dynamics regression (Spearman\,$\rho$\,\%), zero-shot conformational switch (Spearman\,$\rho$\,\%), and remote homology (macro\,$F_1$\,\%), via a shared two-layer MLP probe with best-of-12 selection. \textbf{Bold}/\underline{underline}: best/second-best per row.}
  \vspace{2pt}
  \label{tab:results_str}
  \ifdefined\arxivmode\renewcommand{\arraystretch}{1.08}\footnotesize\else\renewcommand{\arraystretch}{0.95}\scriptsize\fi
  \ifdefined\arxivmode\setlength{\tabcolsep}{1.5pt}\else\setlength{\tabcolsep}{3pt}\fi
  \ifdefined\arxivmode\newcolumntype{T}{l@{\hspace{8pt}}}\else\newcolumntype{T}{l}\fi
  \begin{tabularx}{\textwidth}{Tl *{5}{>{\centering\arraybackslash}X}}
  \toprule
  \textbf{Task} & \textbf{Split} & {\textbf{ProSST}} & {\textbf{ESM3}} & {\textbf{AminoAseed}} & {\textbf{DPLM-2}} & {\textbf{\protok{str}}\ifdefined\arxivmode\newline\else\,\fi{\textbf{(ours)}}} \\
  \midrule[\heavyrulewidth]
  \multicolumn{7}{c}{\textsc{Functional Site Prediction}\; (AUROC\,\%)} \\
  \multirow{2}{*}{InterPro Bind.}
    & Fold   & \bfseries 52.65 & 44.82 & 47.29 & 47.54 & \underline{49.74} \\
    & SupFam & \underline{91.04} & 89.33 & 90.62 & 90.79 & \bfseries 94.53 \\
  \multirow{2}{*}{InterPro Act.}
    & Fold   & 58.99 & 59.70 & 62.13 & \underline{63.38} & \bfseries 64.80 \\
    & SupFam & 90.79 & 88.28 & \underline{91.86} & 90.53 & \bfseries 96.51 \\
  \multirow{2}{*}{InterPro Cons.}
    & Fold   & 55.07 & 55.55 & 57.09 & \bfseries 59.82 & \underline{58.34} \\
    & SupFam & 81.02 & 79.24 & \underline{86.61} & 82.81 & \bfseries 93.67 \\
  \multirow{2}{*}{InterPro Rep.}
    & Fold   & \underline{76.10} & 75.92 & 75.53 & \bfseries 76.30 & 74.81 \\
    & SupFam & 82.53 & 82.07 & \bfseries 85.89 & 81.50 & \underline{84.76} \\
  \multirow{2}{*}{BioLIP2 Bind.}
    & Fold   & 62.71 & 63.16 & 66.54 & \bfseries 80.98 & \underline{78.76} \\
    & SupFam & 66.68 & 65.90 & 68.89 & \underline{82.04} & \bfseries 83.52 \\
  \multirow{2}{*}{BioLIP2 Cat.}
    & Fold   & 63.38 & 65.26 & 69.85 & \bfseries 88.37 & \underline{85.79} \\
    & SupFam & 76.59 & 74.79 & 86.05 & \underline{93.28} & \bfseries 97.70 \\
  ProteinShake
    & Test   & 64.10 & 66.68 & 69.73 & \underline{81.83} & \bfseries 83.77 \\
  \multirow{2}{*}{ProteinGLUE}
    & Fold   & 62.00 & 61.99 & 64.37 & \underline{65.35} & \bfseries 69.75 \\
    & SupFam & 71.86 & 71.70 & 74.52 & \underline{74.63} & \bfseries 81.24 \\
  \cmidrule{1-7}
  \multicolumn{2}{c}{\textit{Average}}
    & 70.37 & 69.63 & 73.13 & \underline{77.28} & \bfseries 79.85 \\
  \midrule[\heavyrulewidth]
  \multicolumn{7}{c}{\textsc{Flexibility \& Dynamics}\; (Spearman\;$\rho$\,\%)} \\
  \multirow{2}{*}{ATLAS B-factor}
    & Fold   & \underline{26.26} & 23.52 & 23.54 & \underline{26.26} & \bfseries 27.90 \\
    & SupFam & \underline{26.53} & 25.68 & 24.95 & 26.01 & \bfseries 29.39 \\
  \multirow{2}{*}{ATLAS RMSF}
    & Fold   & \underline{50.42} & 44.28 & 45.54 & 48.69 & \bfseries 52.17 \\
    & SupFam & \underline{46.90} & 40.76 & 41.13 & 43.06 & \bfseries 47.18 \\
  \multirow{2}{*}{ATLAS N\textsubscript{eq}}
    & Fold   & \underline{55.76} & 45.67 & 49.44 & 46.60 & \bfseries 59.69 \\
    & SupFam & \underline{53.42} & 45.98 & 49.51 & 45.19 & \bfseries 58.45 \\
  \cmidrule{1-7}
  \multicolumn{2}{c}{\textit{Average}}
    & \underline{43.22} & 37.65 & 39.02 & 39.30 & \bfseries 45.80 \\
  \midrule[\heavyrulewidth]
  \multicolumn{7}{c}{\textsc{Conformational Switch}\; (Spearman\;$\rho$\,\%,\; zero-shot)} \\
  \multirow{2}{*}{TM-score}
    & Fold-sw.  & 59.70 & 68.51 & \underline{78.29} & 63.12 & \bfseries 79.27 \\
    & Apo-holo  & 42.17 & 53.48 & \underline{56.02} & 48.68 & \bfseries 57.69 \\
  \multirow{2}{*}{neg-RMSD}
    & Fold-sw.  & 12.98 & 26.70 & \bfseries 40.44 & 17.74 & \underline{38.50} \\
    & Apo-holo  & 2.44 & 16.83 & \underline{22.99} & 10.46 & \bfseries 23.93 \\
  \cmidrule{1-7}
  \multicolumn{2}{c}{\textit{Average}}
    & 29.32 & 41.38 & \underline{49.44} & 35.00 & \bfseries 49.85 \\
  \midrule[\heavyrulewidth]
  \multicolumn{7}{c}{\textsc{Remote Homology}\; (Macro $F_1$\,\%)} \\
  \multirow{3}{*}{TAPE}
    & Fold   & 26.70 & 34.61 & \underline{43.58} & 26.67 & \bfseries 53.48 \\
    & SupFam & 30.59 & 30.07 & \underline{57.54} & 19.32 & \bfseries 72.59 \\
    & Family & 68.92 & 65.42 & \underline{87.44} & 49.49 & \bfseries 96.35 \\
  \cmidrule{1-7}
  \multicolumn{2}{c}{\textit{Average}}
    & 42.07 & 43.37 & \underline{62.85} & 31.83 & \bfseries 74.14 \\
  \bottomrule
  \end{tabularx}
  \ifdefined\arxivmode\else\vspace{-10pt}\fi
  \end{table*}
\section{Results}
\label{sec:results}

\subsection{Representation, Reconstruction, and Generation}
\label{sec:structural_eval}
\paragraph{ProTok is the most consistent structure tokenizer across tasks.}
Table~\ref{tab:results_str} compares \protok{str} with four other discrete structure tokenizers on StructTokenBench.
\protok{str} achieves the highest category average across all four task categories: functional site prediction (leading DPLM-2 by 2.6\% AUROC), flexibility and dynamics (leading ProSST by 2.6\% Spearman~$\rho$), zero-shot conformational discrimination (narrowly edging AminoAseed by 0.4\% Spearman~$\rho$), and remote homology (leading AminoAseed by 11.3\% macro $F_1$).
No other tokenizer achieves top-two performance across all four categories: AminoAseed trails on flexibility and functional site prediction, and DPLM-2 on remote homology and conformational discrimination.
Aggregated across all 28 evaluation splits, \protok{str}'s mean rank of 1.36 is 1.5 positions ahead of the next-best tokenizer (AminoAseed, 2.86), and it places in the top two on 27 of 28 splits.
Section~\ref{sec:ablation} validates that these gains partially arise from multi-modal training via encoder group ablation, and further ablates our orthogonal regularization.

\paragraph{ProTok supports both downstream prediction and high-fidelity reconstruction.}
Few structure tokenizers achieve strong downstream performance and faithful reconstruction together: ESM3 reconstructs with near-perfect fidelity (Table~\ref{tab:tokenizer_summary}) but ranks last on the downstream tasks in Table~\ref{tab:results_str}, while AminoAseed is the opposite.
\protok{str} breaks the pattern, remaining competitive across downstream tasks while recovering structure to 0.939 lDDT on CATH~4.2, within 0.047 of ESM3's native reconstruction.

\paragraph{ProTok tokens are a designable generative substrate.}
A masked generative prior over \protok{str-seq} tokens jointly generates a backbone and its amino acid sequence by iterative unmasking, predicting the per-group product codes at each position (Appendix~\ref{app:generation}).
We assess generated backbones by self-consistency: a backbone is designable when a redesigned sequence refolds to it with self-consistency TM-score (scTM)~$\geq$~0.5.
At a parameter budget matched to GeoBPE's generative model, 52\% of the backbones sampled from ProTok's prior are designable, compared with 21\% for GeoBPE.
The samples are co-designable under the prior's own sequence 18\% of the time and remain largely unique; Appendix~\ref{app:generation} (Table~\ref{tab:generation_full}) reports full metrics, together with a larger 156M prior that performs comparably.
ESM3 and DPLM-2 reach higher unconditional designability, but with generators an order of magnitude larger trained jointly on far more data; because the GeoBPE comparison holds generator size fixed, the gap between the two reflects the tokens themselves.

\paragraph{ProTok's codebooks are broadly utilized at high capacity.}
\protok{str} is the only tokenizer in our comparison to combine high codebook utilization and high normalized perplexity while encoding 64 bits per residue, several times any flat vocabulary considered; product quantization splits each residue across small, individually saturable codebooks, and an orthogonal regularizer keeps the SimVQ projections well-conditioned (Section~\ref{sec:ablation}).
Appendix~\ref{app:codebook_analysis} (Table~\ref{tab:tokenizer_summary}) reports per-tokenizer utilization, perplexity, and reconstruction.

\subsection{Codebook Probing}
\label{sec:crossmodal}

\paragraph{ProTok's structure tokens lead matched-probe decoding.}
Matched probes that isolate token information content from each tokenizer's decoder capacity place \protok{str}'s structure tokens highest on structure-prediction lDDT (0.724) and near the top on inverse-folding recovery, close behind DPLM-2 (Table~\ref{tab:if_sp}).

\paragraph{Structure tokens retrieve function zero-shot.}
Ranking proteins by cosine similarity over mean-pooled structure tokens, \protok{str} leads every structure tokenizer on all nine label hierarchies (Table~\ref{tab:retrieval}).
The margin grows from 0.012 P@1 at coarse CATH class to 0.185 at EC level~1; since EC and GO labels track chemistry and biological role rather than backbone geometry, these margins suggest \protok{str}'s codes carry signal beyond backbone reconstruction.

\begin{table*}[t]
\centering
\caption{\textbf{Zero-shot functional retrieval on CATH~4.2.}
Top-1 precision (P@1) from mean-pooled structure tokens ranked by cosine similarity, across nine label hierarchies in three families (CATH, Enzyme Commission, Gene Ontology); a query and candidate match when their label sets intersect. \textbf{Bold}/\underline{underline}: best/second-best per column.}
\vspace{2pt}
\label{tab:retrieval}
\renewcommand{\arraystretch}{1.05}
\footnotesize
\begin{tabularx}{\textwidth}{l *{9}{>{\centering\arraybackslash}X}}
\toprule
& \multicolumn{3}{c}{\textbf{CATH}} & \multicolumn{3}{c}{\textbf{EC}} & \multicolumn{3}{c}{\textbf{GO}} \\
\cmidrule(lr){2-4} \cmidrule(lr){5-7} \cmidrule(lr){8-10}
\textbf{Tokenizer} & \textbf{Class} & \textbf{Arch.} & \textbf{Topo.} & \textbf{EC-1} & \textbf{EC-2} & \textbf{EC-3} & \textbf{MF} & \textbf{BP} & \textbf{CC} \\
\midrule[\heavyrulewidth]
ProSST       & \underline{0.933} & \underline{0.757} & \underline{0.655} & \underline{0.510} & 0.345            & 0.337             & 0.355            & \underline{0.269} & \underline{0.407} \\
ESM3         & 0.924             & 0.740             & 0.634             & 0.498             & \underline{0.390} & \underline{0.361} & 0.359            & 0.244             & 0.350 \\
AminoAseed   & 0.888             & 0.701             & 0.632             & 0.506             & 0.373             & 0.321             & \underline{0.379} & \underline{0.269} & 0.350 \\
DPLM-2       & 0.888             & 0.652             & 0.541             & 0.470             & 0.321             & 0.285             & 0.309            & 0.224             & 0.394 \\
\midrule
\protok{str} & \textbf{0.945}    & \textbf{0.868}    & \textbf{0.817}    & \textbf{0.695}    & \textbf{0.562}    & \textbf{0.502}    & \textbf{0.490}   & \textbf{0.407}    & \textbf{0.516} \\
\bottomrule
\end{tabularx}
\end{table*}

\paragraph{Codebook groups carry complementary information.}
Activating additional codebook groups at inference, across three nested modes (\protok{str}, \protok{str-seq}, \protok{str-seq-fun}), exercises the any-subset capability on five StructTokenBench tasks (Appendix~\ref{app:probing_extra}, Table~\ref{tab:complementarity}).
On all four functional-site tasks performance improves monotonically with each added group; on catalytic-site classification, fold-level AUROC rises from 0.858 (\protok{str}) to 0.916 (\protok{str-seq}) to 0.931 (\protok{str-seq-fun}).
On ATLAS RMSF flexibility regression, where labels reflect local backbone dynamics rather than protein function, \protok{str-seq} attains the best Spearman $\rho$ on both holdout splits while \protok{str-seq-fun} regresses, as the protein-level function tokens add noise to a residue-level structural prediction task.
The groups thus encode complementary information, with gains appearing when the added groups correlate with the task's labels.

\subsection{Ablation Study}
\label{sec:ablation}

Table~\ref{tab:ablation} ablates ProTok's encoder groups and orthogonal regularizer.
The top section adds encoder groups cumulatively during training while restricting downstream inference to the structure codebooks, so any gain reflects information that joint training has distilled into them.
Adding the local geometry, sequence, and function groups in turn raises split-averaged remote homology macro $F_1$ from \textrm{64.1}\% (global structure only) to \textrm{74.5}\% (full), validating modality dropout as a mechanism for transferring complementary signal into structure tokens.
Removing the orthogonal regularizer from the full model collapses the average effective rank of the SimVQ projections from \textrm{80.4}\% to \textrm{33.3}\% and pulls remote homology classification performance back to the level of the structure-only baseline, indicating that codebook diversity is essential for global structural discrimination.

%% file: content/06_future.tex
\section{Discussion and Conclusion}
\label{sec:conclusion}

\subsection{Discussion}

Modality dropout imposes an objective distinct from multi-task learning or contrastive alignment: at each step we sample one of $\textrm{2}^G - \textrm{1} = \textrm{15}$ non-empty subsets of the $G = \textrm{4}$ codebook groups to keep active
\begin{table}[t]
\centering
\footnotesize
\setlength{\tabcolsep}{5pt}
\caption{\textbf{Ablation study.}
We ablate ProTok's encoder groups (top, cumulative) and the orthogonal regularizer (bottom).
We report split-averaged TAPE macro $F_1$ (\%), ProteinGLUE AUROC (\%), and SimVQ projection effective rank (\%). Each variant is trained separately under an identical schedule, so the full row differs slightly from Table~\ref{tab:results_str}.}
\vspace{-4pt}
\label{tab:ablation}
\begin{tabular}{l c c c}
\toprule
\textbf{Variant} & \textbf{TAPE} & \textbf{ProteinGLUE} & \textbf{Rank} \\
\midrule
Global structure  & \textrm{64.1} & \textrm{68.2} & --- \\
\raisebox{0.3ex}{\tiny$+$} Local geom.     & \textrm{71.1} & \textrm{73.6} & --- \\
\raisebox{0.3ex}{\tiny$+$} Sequence        & \textrm{70.4} & \textrm{74.9} & --- \\
\raisebox{0.3ex}{\tiny$+$} Function (full) & \textrm{74.5} & \textrm{73.3} & \textrm{80.4} \\
\midrule
\raisebox{0.3ex}{\tiny$-$} Ortho. reg.   & \textrm{64.1} & \textrm{72.6} & \textrm{33.3} \\
\bottomrule
\end{tabular}
\end{table}
and require those active codes to support reconstruction on their own.
We attribute the cross-modal transfer in Section~\ref{sec:crossmodal} to this objective, which also yields single-checkpoint any-subset inference at test time.
It also addresses the reconstruction-versus-representation trade-off of prior tokenizers (Section~\ref{sec:structural_eval}): ProTok's four 256-entry codebooks per group supply roughly $\textrm{4}\times\textrm{10}^\textrm{9}$ compositions at each residue, enough for high-fidelity reconstruction, while modality dropout distills sequence and functional semantics into the same codes.

ProTok is a tokenizer, not a task-specific predictor: our results show that one vocabulary can serve representation, reconstruction, and generation, not that it surpasses specialized function predictors.

\subsection{Limitations}
Our 311K-chain corpus trains only the quantizer, fusion module, and decoders, since the encoders are frozen, but it remains orders of magnitude smaller than the corpora used to train production protein language models; longer training on more diverse structural data at PLM scale (for instance, AlphaFold Database~\citep{varadi2022alphafold} clusters) and broader functional annotation may yield further gains.
Our generation results are preliminary: the prior is small, trained on short CATH chains (at most 128 residues) and evaluated on 50 samples.
Scaling a ProTok-native autoregressive or masked-diffusion language model to longer proteins and larger corpora, with conditional generation, is the main direction for future work.
Adding modality groups is also not universally beneficial: the protein-level function group helps residue-level functional-site prediction but can hurt dynamics regression, as on ATLAS RMSF.

\subsection{Conclusion}
ProTok shows that multi-codebook quantization with modality dropout yields tokens that both reconstruct geometry faithfully and carry sequence and functional signal.
The resulting vocabulary places among the top two structure-only tokenizers on 27 of 28 StructTokenBench splits and supports any subset of modalities at inference from a single checkpoint, collapsing a trade-off that prior structure tokenizers have left unresolved; a prior over its tokens further co-generates backbones and sequences, over half of them designable.
By pushing multi-modal supervision into the tokenizer, ProTok supplies a stronger substrate for the next generation of protein language models, and for the protein design and functional prediction tools that increasingly depend on them.

%% file: content/appendix.tex
\appendix

\renewcommand{\thefigure}{S\arabic{figure}}
\renewcommand{\thetable}{S\arabic{table}}
\setcounter{figure}{0}
\setcounter{table}{0}

\section{Training Details}
\label{app:training_details}

\paragraph{Structure reconstruction targets.}
We adopt the structure reconstruction objective of ESM3~\citep{hayes2024simulating} and AminoAseed~\citep{guo2024structtokenbench} without modification; the Stage~1 structure decoder produces five outputs (Section~\ref{sec:decoder}), each supervised by a distinct loss:
\begin{itemize}
    \item Pairwise backbone-atom distances and backbone direction vectors, both supervised with mean squared error;
    \item Binned $\text{C}_\beta$ distances, supervised with cross-entropy over 64 quadratically-spaced bins;
    \item Binned pairwise backbone-direction cosines, supervised with cross-entropy over 16 linearly-spaced bins;
    \item Amino acid identity (inverse folding), supervised with cross-entropy over the 20 canonical amino acids.
\end{itemize}
On AlphaFold-predicted chains, every reconstruction term is weighted per-residue by the pLDDT confidence; PDB chains use uniform weighting.

\paragraph{Loss coefficients.}
In Equation~\ref{eq:total_loss}, $\loss_\mathrm{recon}$, $\loss_\mathrm{VQ}$, $\loss_\mathrm{seq}$, and $\loss_\mathrm{func}$ enter with unit weight; we set $\lambda_\mathrm{ortho} = \textrm{0.01}$ and $\lambda_\mathrm{entropy} = \textrm{0.1}$.
Inside $\loss_\mathrm{VQ}$, the commitment and codebook alignment terms use $\lambda_\mathrm{commit} = \textrm{0.25}$ and $\lambda_\mathrm{quant} = \textrm{1}$.
$\loss_\mathrm{func}$ is a symmetric InfoNCE loss~\citep{oord2018representation} with temperature $\tau = \textrm{0.1}$; proteins without UniProt annotations are excluded from its minibatch.
$\loss_\mathrm{entropy}$ is the gap between the per-sample prediction entropy and the per-codebook usage entropy, encouraging confident per-residue assignments together with uniform code utilization across the dataset.

\paragraph{Stage~2 decoder fine-tuning.}
Stage~2 fine-tunes a 30-layer ESM3-pretrained structure token decoder on Stage~1's frozen structure and local geometry token streams, with the sequence and function modalities disabled and the encoder, structure codebook, and local geometry codebook held fixed.
The Stage~2 decoder retains the five Stage~1 reconstruction targets and adds two auxiliary heads that regress the pLDDT and predicted aligned error (PAE) confidence signals of AlphaFold~2 from the decoder's hidden states; each auxiliary loss contributes to the total objective with weight $\textrm{0.01}$.

\paragraph{Optimization.}
Both stages train for 25 epochs on 8 NVIDIA B200 GPUs at effective batch size 256, optimized with AdamW~\citep{loshchilov2017decoupled} at learning rate $\textrm{5} \times \textrm{10}^{-\textrm{4}}$, weight decay $\textrm{0.01}$, and a cosine schedule with 2{,}000 warmup steps.
Training runs in \texttt{bfloat16} mixed precision with gradient clipping at norm $\textrm{1.0}$.
Encoder inputs receive $\sigma = \textrm{0.1}$\,\AA\ isotropic Gaussian coordinate noise during training; reconstruction targets use clean coordinates.
A weighted random sampler produces balanced per-batch Protein Data Bank and Swiss-Prot ratios regardless of the overall corpus composition.
Stage~1 has approximately 146M trainable parameters, while Stage~2 has approximately 632M trainable parameters.

\section{Evaluation Setup}
\label{app:eval_setup}

\paragraph{StructTokenBench benchmarks and hyperparameter sweep.}
StructTokenBench~\citep{guo2024structtokenbench} comprises seven benchmarks containing 14 prediction tasks, each evaluated on one or more held-out splits for a total of 28 test splits.
For the 12 supervised tasks, we attach a two-layer MLP over frozen tokens and train on each task's training split.
For each (tokenizer, task) pair we sweep a 12-configuration grid over learning rate $\{\textrm{10}^{-\textrm{3}}, \textrm{10}^{-\textrm{4}}, \textrm{10}^{-\textrm{5}}\}$, dropout $\{\textrm{0.1}, \textrm{0.3}\}$, and hidden size $\{\textrm{256}, \textrm{512}\}$, each trained for 10{,}000 steps of AdamW~\citep{loshchilov2017decoupled} with a cosine schedule, batch size~8, and a shared random seed.
We select the configuration with the best validation metric ($F_1$ for classification, Spearman~$\rho$ for regression) and report test metrics on every split of that task from the selected run.
TAPE~\citep{rao2019evaluating} uses family-, superfamily-, and fold-holdout splits; BioLIP2~\citep{zhang2024biolip2}, InterPro~\citep{paysanlafosse2023interpro}, ATLAS~\citep{vandermeersche2024atlas}, and ProteinGLUE~\citep{capel2022proteinglue} use fold- and superfamily-holdout splits; ProteinShake~\citep{kucera2023proteinshake} uses a single test split.
The two Conformational Switch tasks (TM-score~\citep{zhang2004scoring} and negative RMSD) are evaluated zero-shot on fold-switching and apo-holo splits via Needleman-Wunsch~\citep{needleman1970general} alignment over structure tokens, sweeping the softmax temperature over $\{\textrm{5}, \textrm{10}, \textrm{25}, \textrm{50}, \textrm{100}\}$ and reporting the best Spearman correlation with the ground truth labels.

\paragraph{Seed variability.}
\label{app:seeds}
To quantify run-to-run variance, we rerun the validation-selected configuration of \protok{str} with five seeds (1234, 42, 7, 2024, 31337) on five representative tasks spanning global classification, residue-level classification, and residue-level regression (Table~\ref{tab:seeds}).
The seed standard deviation is at most 2.7 macro $F_1$ (remote homology fold) and at most 0.4 AUROC or Spearman~$\rho$ on the other tasks, small relative to the gap to the best baseline on all but ATLAS RMSF superfamily, where the two are within seed variance; on the one split in this table where a baseline leads (BioLIP2 catalytic fold, DPLM-2), it leads across all seeds.

\begin{table}[h]
\centering
\footnotesize
\setlength{\tabcolsep}{8pt}
\caption{\textbf{Seed variability of \protok{str}.} Mean $\pm$ standard deviation over five seeds of the validation-selected probe configuration, with the single-run value reported in Table~\ref{tab:results_str} and the strongest baseline on each split.}
\vspace{2pt}
\label{tab:seeds}
\begin{tabular}{llccl}
\toprule
\textbf{Task} & \textbf{Split} & \textbf{\protok{str} (5 seeds)} & \textbf{\protok{str} (Table 1)} & \textbf{Best baseline} \\
\midrule[\heavyrulewidth]
\multicolumn{5}{c}{\textsc{Functional Site Prediction}\; (AUROC\,\%)} \\[2pt]
\multirow{2}{*}{BioLIP2 Cat.} & Fold   & 86.8 $\pm$ 0.2 & 85.79 & 88.37 (DPLM-2) \\
                              & SupFam & 97.8 $\pm$ 0.1 & 97.70 & 93.28 (DPLM-2) \\
ProteinShake                  & Test   & 84.3 $\pm$ 0.4 & 83.77 & 81.83 (DPLM-2) \\
\multirow{2}{*}{ProteinGLUE}  & Fold   & 69.5 $\pm$ 0.4 & 69.75 & 65.35 (DPLM-2) \\
                              & SupFam & 80.9 $\pm$ 0.3 & 81.24 & 74.63 (DPLM-2) \\
\midrule[\heavyrulewidth]
\multicolumn{5}{c}{\textsc{Flexibility \& Dynamics}\; (Spearman\;$\rho$\,\%)} \\[2pt]
\multirow{2}{*}{ATLAS RMSF}   & Fold   & 52.7 $\pm$ 0.3 & 52.17 & 50.42 (ProSST) \\
                              & SupFam & 48.0 $\pm$ 0.3 & 47.18 & 46.90 (ProSST) \\
\midrule[\heavyrulewidth]
\multicolumn{5}{c}{\textsc{Remote Homology}\; (Macro $F_1$\,\%)} \\[2pt]
\multirow{3}{*}{TAPE}         & Fold   & 51.7 $\pm$ 2.7 & 53.48 & 43.58 (AminoAseed) \\
                              & SupFam & 69.8 $\pm$ 2.3 & 72.59 & 57.54 (AminoAseed) \\
                              & Family & 96.0 $\pm$ 1.0 & 96.35 & 87.44 (AminoAseed) \\
\bottomrule
\end{tabular}
\end{table}

\paragraph{Native reconstruction quality.}
We measure native reconstruction quality on the 1{,}120 test chains of CATH~4.2~\citep{sillitoe2021cath} in Table~\ref{tab:tokenizer_summary}.
For each baseline tokenizer, the chain is passed through its encoder, quantizer, and pretrained native decoder; for ProTok, the chain is passed through the Stage~1 encoder and quantizer and decoded with the Stage~2 fine-tuned decoder.
We report $\textrm{C}_\alpha$ lDDT~\citep{mariani2013lddt}, following the CASP convention. ProSST~\citep{chen2024prosst} does not have a native decoder, so we exclude it from this analysis.

\paragraph{Cross-modal probes.}
The inverse folding and structure prediction probes (Table~\ref{tab:if_sp}) each train a 4-layer transformer (hidden size 256, 8 attention heads) over frozen structure tokens on the CATH~4.2 training split.
The inverse folding probe predicts amino acid identity at each residue with a linear head and reports top-1 amino acid recovery; the structure prediction probe predicts an SE(3) backbone frame at each residue and reports $\textrm{C}_\alpha$ lDDT.
We train for 10 epochs at batch size 8 with AdamW and a cosine schedule (200 warmup steps), and report the test metric of the checkpoint with the best validation score.
Functional retrieval is entirely zero-shot on the CATH~4.2 test set: for each query, we mean-pool its structure tokens into a single protein-level vector and rank candidates by cosine similarity, reporting top-1 precision (P@1).
A protein matches a query if its label set intersects the query's.
We evaluate on three label hierarchies: the CATH structural hierarchy (Class: 4 classes, Architecture: 19, Topology: 130; 1{,}120 proteins); Enzyme Commission numbers at levels 1--3 (7, 36, 65 classes; 249 annotated enzymes obtained via SIFTS~\citep{velankar2013sifts} and Swiss-Prot); and Gene Ontology~\citep{ashburner2000go,aleksander2023gene} terms from the same source (biological process: 509 proteins, molecular function: 512, cellular component: 477).

\paragraph{Assets and licenses.}
All data and models we use are publicly available, and we use them for non-commercial research under their terms.
Protein Data Bank entries are released under CC0~1.0; AlphaFold Database predictions, UniProt annotations, and CATH are released under CC~BY~4.0; StructTokenBench, ESM-2, ESM-IF1, ESMFold, and ProteinMPNN are released under the MIT license; the ESM3 open model is released under the EvolutionaryScale Cambrian Non-Commercial License; and Qwen3-Embedding is released under Apache~2.0.

\section{Codebook Analysis}
\label{app:codebook_analysis}

This section reports three measurements of \protok{}'s discrete vocabulary on the CATH~4.2 test split (1{,}120 chains): probe-free per-group quantization fidelity, per-modality codebook utilization and perplexity, and cross-group mutual information.
Function group statistics are restricted to the $\textrm{290}$ chains for which functional annotations are available.

\paragraph{Cross-tokenizer utilization and reconstruction.}
Table~\ref{tab:tokenizer_summary} compares \protok{str} against four structure tokenizers on codebook utilization rate (UR), normalized perplexity (PPL), and native reconstruction lDDT.
ESM3 and DPLM-2 reach high UR but with skewed code frequencies; AminoAseed has the lowest UR with similarly imbalanced usage.
ProSST nearly saturates its codebook (UR 0.997, PPL 0.871), but $k$-means clustering populates every centroid by construction, sidestepping the codebook utilization problem rather than solving it through training.
\protok{str} is the only tokenizer to combine high UR and high PPL at high capacity: it encodes 64 bits per residue, several times any flat vocabulary in the comparison, with UR 0.958 and PPL 0.769.

\begin{table}[h]
\centering
\footnotesize
\setlength{\tabcolsep}{8pt}
\caption{\textbf{Structure tokenizer summary.}
We compute codebook utilization rate (UR), normalized perplexity (PPL), and native reconstruction lDDT for \protok{str} and four structure tokenizers on CATH~4.2.}
\vspace{2pt}
\label{tab:tokenizer_summary}
\begin{tabular}{lcccccc}
\toprule
\textbf{Tokenizer} & \textbf{Vocabulary} & \textbf{Dim.} & \textbf{Bits} & \textbf{UR} & \textbf{PPL} & \textbf{lDDT} \\
\midrule[\heavyrulewidth]
ProSST       & 2{,}048          &     256 & 11 & 0.997 & 0.871 & --- \\
ESM3         & 4{,}096          &     128 & 12 & 0.946 & 0.517 & 0.986 \\
AminoAseed   & 512              & 1{,}024 &  9 & 0.807 & 0.485 & 0.634 \\
DPLM-2       & 8{,}192          &     128 & 13 & 0.917 & 0.612 & 0.900 \\
\midrule
\protok{str} & 8\,$\times$\,256 &     256 & 64 & 0.958 & 0.769 & 0.939 \\
\bottomrule
\end{tabular}
\end{table}

\begin{table}[h]
\centering
\footnotesize
\setlength{\tabcolsep}{8pt}
\caption{\textbf{Per-group codebook diagnostics.}
We measure utilization rate (UR), normalized perplexity (PPL), cosine similarity between each per-codebook input $\mathbf{z}$ and its codebook lookup $\mathbf{q}$, and the corresponding relative MSE. Values are averaged across each group's four codebooks. The reference row gives a nearest-of-256 match against random unit vectors.
}
\vspace{2pt}
\label{tab:per_group_diagnostics}
\begin{tabular}{lccccccc}
\toprule
\textbf{Codebook group} & \textbf{Modality} & \textbf{Encoder} & \textbf{Vocabulary} & \textbf{UR} & \textbf{PPL} & \textbf{$\cos(\mathbf{z}, \mathbf{q})$} & \textbf{Rel.\ MSE} \\
\midrule
Global structure & Structure & AminoAseed & \textrm{4}\,$\times$\,\textrm{256} & \textrm{0.926} & \textrm{0.751} & \textrm{0.354} & \textrm{1.291} \\
Local geometry   & Structure & ESM-IF1    & \textrm{4}\,$\times$\,\textrm{256} & \textrm{0.991} & \textrm{0.787} & \textrm{0.287} & \textrm{1.425} \\
Sequence         & Sequence  & ESM-2      & \textrm{4}\,$\times$\,\textrm{256} & \textrm{0.997} & \textrm{0.940} & \textrm{0.276} & \textrm{1.447} \\
Function         & Function  & Qwen3      & \textrm{4}\,$\times$\,\textrm{256} & \textrm{0.396} & \textrm{0.250} & \textrm{0.236} & \textrm{1.528} \\
\midrule
Random reference & --- & --- & --- & --- & --- & \textrm{0.175} & \textrm{1.650} \\
\bottomrule
\end{tabular}
\end{table}

\paragraph{Per-group codebook utilization.}
The first three groups in Table~\ref{tab:per_group_diagnostics} reach broad codebook usage with relatively uniform code distributions (UR between \textrm{0.93} and \textrm{1.00}, PPL between \textrm{0.75} and \textrm{0.94}).
The function group uses fewer than half of its entries (UR \textrm{0.40}, PPL \textrm{0.25}). However, its codebooks operate at protein granularity, so usage is bounded by the \textrm{290} per-protein function annotations rather than the per-residue token count of the other groups.

\paragraph{Per-group quantization fidelity.}
The right columns of Table~\ref{tab:per_group_diagnostics} report the directional information that each per-codebook lookup preserves.
The global structure group has the highest fidelity with a mean cosine similarity of \textrm{0.354} (twice the random floor of \textrm{0.175}), local geometry and sequence are near-identical (\textrm{0.287} and \textrm{0.276}, respectively), and function is lowest (\textrm{0.236}).
We hypothesize that the ordering reflects upstream encoder pretraining: the global structure encoder originates from AminoAseed, itself a VQ tokenizer, and produces features that already cluster around discrete prototypes, whereas the local geometry, sequence, and function groups quantize representations from encoders trained on continuous objectives (ESM-IF1, ESM-2, Qwen3) whose smoother manifolds align less tightly with any fixed codebook.

\paragraph{Cross-group mutual information.}
\begin{figure}[h]
\centering
\includegraphics[width=0.55\textwidth]{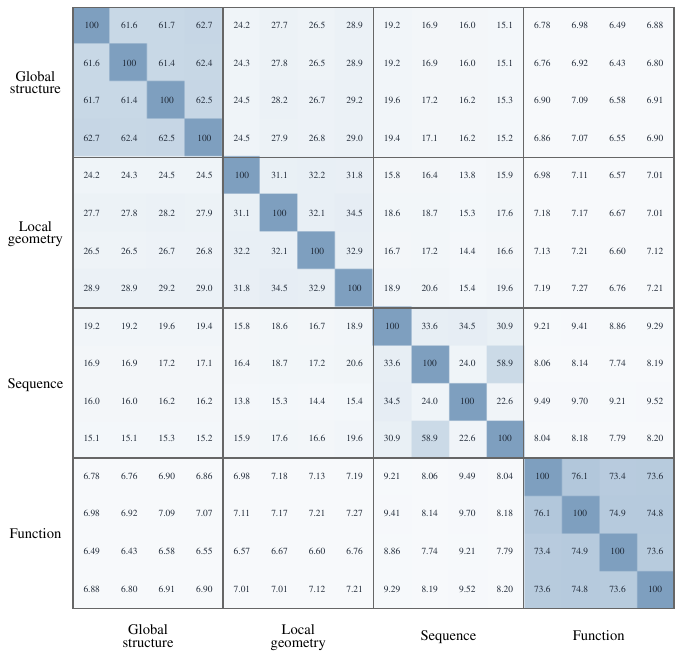}
\caption{\textbf{Cross-group mutual information.}
We compute NMI~\% between the \textrm{16} discrete token streams (\textrm{4} codebooks per group); cells involving the function group use only the \textrm{290}-chain annotated subset. Darker borders separate the four groups.
}
\label{fig:codebook_group_nmi}
\end{figure}
Figure~\ref{fig:codebook_group_nmi} reports normalized mutual information (NMI) between the four codebook groups.
Within-group NMI tracks input granularity: the protein-level function group is most redundant, global structure is intermediate, and the per-residue local geometry and sequence groups are least redundant.
Cross-group NMI is uniformly small: global structure and local geometry share the most information, while function is essentially independent of every other group.
This decoupling indicates that modality dropout pushes each group to encode complementary signal at training time (though low NMI between discrete vocabularies is a necessary rather than sufficient condition for complementarity).

\section{Sequence Token Probing}
\label{app:seq_probe_trim}

Table~\ref{tab:results_seq_trim} compares three sequence-only representations on four StructTokenBench tasks: an amino acid baseline (a frozen $\textrm{21}\times\textrm{1024}$ random codebook indexed by amino acid identity), \protok{seq}, and continuous ESM-2 \textrm{650}M.
All three use the same \textrm{12}-configuration hyperparameter sweep and selection protocol as Table~\ref{tab:results_str}.
\protok{seq} beats the amino acid baseline on every task, motivating learned sequence codes over a fixed amino acid codebook. It trails ESM-2 by $\textrm{1.3}$ to $\textrm{8.5}$ points across the four tasks, suggesting that the discrete bottleneck preserves most but not all of the signal in continuous PLM hidden states.

\begin{table}[h]
\centering
\caption{\textbf{Quality of sequence-only representations.} We compare three sequence-only representations on four StructTokenBench tasks, reporting AUROC\,\% for functional sites and macro\,$F_1$\,\% for remote homology. \textbf{Bold} marks the best per column.}
\vspace{2pt}
\label{tab:results_seq_trim}
\setlength{\tabcolsep}{4pt}
\renewcommand{\arraystretch}{1.0}
\footnotesize
\begin{tabularx}{\textwidth}{l *{4}{>{\centering\arraybackslash\hsize=.94\hsize}X} >{\centering\arraybackslash\hsize=1.42\hsize}X *{3}{>{\centering\arraybackslash\hsize=.94\hsize}X}}
\toprule
& \multicolumn{2}{c}{\textbf{BioLIP2 Cat.}}
  & \multicolumn{2}{c}{\textbf{InterPro Act.}}
  & \textbf{ProteinShake}
  & \multicolumn{3}{c}{\textbf{TAPE}} \\
\cmidrule(lr){2-3} \cmidrule(lr){4-5} \cmidrule(lr){6-6} \cmidrule(lr){7-9}
\textbf{Tokenizer}
  & Fold & SupFam & Fold & SupFam & Test & Fold & SupFam & Family \\
\midrule
Amino acid   & \textrm{65.23}          & \textrm{72.56}          & \textrm{53.64}          & \textrm{74.68}          & \textrm{59.72}          & \textrm{9.39}           & \textrm{6.27}           & \textrm{25.64} \\
ESM-2        & \textbf{\textrm{93.11}} & \textbf{\textrm{97.76}} & \textbf{\textrm{68.66}} & \textbf{\textrm{97.30}} & \textbf{\textrm{88.31}} & \textbf{\textrm{46.14}} & \textbf{\textrm{75.98}} & \textbf{\textrm{98.72}} \\
\midrule
\protok{seq} & \textrm{90.56}          & \textrm{97.66}          & \textrm{61.02}          & \textrm{96.99}          & \textrm{79.79}          & \textrm{40.46}          & \textrm{66.31}          & \textrm{97.16} \\
\bottomrule
\end{tabularx}
\end{table}

\section{Generation}
\label{app:generation}

We train a masked generative prior over \protok{str-seq} tokens (12 codes per residue) on a CATH~4.2 corpus restricted to chains of at most 128 residues, matching the regime in which GeoBPE reports generation.
The prior is a bidirectional transformer trained to recover masked residues; it samples by iteratively unmasking residue positions and, at each committed position, samples the per-group product codes jointly, so decoding operates on a length-$L$ stream rather than a flattened $12L$ sequence.
We draw 50 samples, decode each token stream to a backbone and amino acid sequence, and evaluate self-consistency: each backbone is redesigned with ProteinMPNN, the designed sequences are refolded with ESMFold, and the refolded structure is compared to the generated backbone by TM-score (scTM).
A backbone is \emph{designable} when its best scTM reaches 0.5 and \emph{co-designable} when the prior's own generated sequence refolds to scTM~$\geq$~0.5.
Following GeoBPE~\citep{sun2026geobpe}, diversity is one minus the mean pairwise TM-score among generated backbones (higher is more diverse), and uniqueness is the fraction of samples retained by a greedy pass that keeps a sample only if its TM-score to every retained sample is below 0.5.
Table~\ref{tab:generation_full} reports designability, scTM, diversity, and uniqueness for a parameter-matched 75M prior, a larger 156M prior, and GeoBPE.
The 75M prior is also co-designable 18\% of the time and the 156M prior 20\%; GeoBPE does not report co-designability.
The 75M prior is matched to GeoBPE's generative model size, removing the parameter-count confound; the 156M prior performs comparably, indicating the matched-budget result is not limited by prior size.
The two systems use different folding oracles (ESMFold here, OmegaFold in GeoBPE's evaluation), so this is a system-level comparison rather than a controlled oracle match.

\begin{table}[h]
\centering
\footnotesize
\setlength{\tabcolsep}{8pt}
\caption{\textbf{Co-generation on CATH~4.2.}
Designability (fraction of samples with self-consistency scTM~$\geq$~0.5 under a redesigned sequence), mean scTM, diversity, and uniqueness for the two ProTok priors and GeoBPE's generative model at a matched parameter budget.
GeoBPE values are as reported in its paper under a different folding oracle; co-designability, which GeoBPE does not report, is given in the text.}
\vspace{2pt}
\label{tab:generation_full}
\begin{tabular}{lcccc}
\toprule
\textbf{Prior} & \textbf{Designable\,\%} & \textbf{scTM} & \textbf{Diversity} & \textbf{Unique\,\%} \\
\midrule[\heavyrulewidth]
GeoBPE                  & 21.0 & 0.405 & 0.731 & 76.0 \\
\midrule
\protok{str-seq} (75M)  & 52.0 & 0.538 & 0.713 & 80.0 \\
\protok{str-seq} (156M) & 54.0 & 0.530 & 0.705 & 76.0 \\
\bottomrule
\end{tabular}
\end{table}

\section{Matched-Probe and Complementarity Tables}
\label{app:probing_extra}

This appendix reports the full codebook-group complementarity and token-probe tables summarized in Section~\ref{sec:crossmodal}; the zero-shot functional retrieval table appears in the main text (Table~\ref{tab:retrieval}).

\newcommand{\tokenprobetable}{%
\begin{table}[h]
\centering
\footnotesize
\setlength{\tabcolsep}{8pt}
\caption{\textbf{Standardized token probes on CATH~4.2.}
We train probes on each tokenizer's structure tokens. Inverse folding reports amino acid recovery (AAR); structure prediction reports per-residue lDDT.}
\vspace{2pt}
\label{tab:if_sp}
\begin{tabular}{llcc}
\toprule
\textbf{Tokenizer} & \textbf{Training} & \textbf{AAR} & \textbf{lDDT} \\
\midrule[\heavyrulewidth]
ProSST       & Structure-only & 0.143 & 0.636 \\
ESM3         & Structure-only & 0.251 & 0.719 \\
AminoAseed   & Structure-only & 0.191 & 0.694 \\
DPLM-2       & Joint          & 0.369 & 0.692 \\
\midrule
\protok{str} & Joint          & 0.359 & 0.724 \\
\bottomrule
\end{tabular}
\end{table}}
\ifdefined\arxivmode\tokenprobetable\fi

\begin{table}[h]
\centering
\caption{\textbf{Codebook group complementarity.}
We evaluate three nested ProTok inference modes on five StructTokenBench tasks, following the probing setup of Table~\ref{tab:results_str} and varying only the number of input codebooks across modes.
We report AUROC~\% for functional sites and Spearman $\rho$~\% for ATLAS RMSF.
\textbf{Bold} marks the best mode per column.}
\vspace{2pt}
\label{tab:complementarity}
\renewcommand{\arraystretch}{1.05}
\footnotesize
\setlength{\tabcolsep}{3pt}
\begin{tabular*}{\textwidth}{@{\extracolsep{\fill}}l c *{10}{c} @{}}
\toprule
& & \multicolumn{2}{c}{\textbf{BioLIP2 Bind.}}
  & \multicolumn{2}{c}{\textbf{BioLIP2 Cat.}}
  & \multicolumn{2}{c}{\textbf{InterPro Bind.}}
  & \multicolumn{2}{c}{\textbf{InterPro Act.}}
  & \multicolumn{2}{c}{\textbf{ATLAS RMSF}} \\
\cmidrule(lr){3-4} \cmidrule(lr){5-6} \cmidrule(lr){7-8} \cmidrule(lr){9-10} \cmidrule(lr){11-12}
\textbf{Mode} & \ifdefined\arxivmode\textbf{No.\,CB}\else\textbf{Codebooks}\fi
  & Fold & SupFam & Fold & SupFam & Fold & SupFam & Fold & SupFam & Fold & SupFam \\
\midrule[\heavyrulewidth]
\protok{str}     & \textrm{8}  & \textrm{78.8} & \textrm{83.5} & \textrm{85.8} & \textrm{97.7} & \textrm{49.7} & \textrm{94.5} & \textrm{64.8} & \textrm{96.5} & \textrm{52.2} & \textrm{47.2} \\
\protok{str-seq}     & \textrm{12} & \textrm{83.3} & \textrm{87.0} & \textrm{91.6} & \textrm{98.6} & \textrm{51.4} & \textrm{96.1} & \textrm{65.5} & \textrm{96.8} & \textbf{\textrm{53.4}} & \textbf{\textrm{49.4}} \\
\protok{str-seq-fun} & \textrm{16} & \textbf{\textrm{83.8}} & \textbf{\textrm{88.1}} & \textbf{\textrm{93.1}} & \textbf{\textrm{98.7}} & \textbf{\textrm{56.2}} & \textbf{\textrm{97.2}} & \textbf{\textrm{67.1}} & \textbf{\textrm{98.1}} & \textrm{48.0} & \textrm{39.7} \\
\bottomrule
\end{tabular*}
\end{table}

\ifdefined\arxivmode\FloatBarrier\else\tokenprobetable\fi

\paragraph{Matched-probe decoding.}
Two matched probes on CATH~4.2, amino acid recovery (AAR) from inverse folding and per-residue lDDT from structure prediction, isolate token information content from each tokenizer's decoder capacity (Table~\ref{tab:if_sp}).
\protok{str} attains the highest structure-prediction lDDT (0.724), surpassing every baseline, and reaches 0.359 AAR on inverse folding, near-tying DPLM-2 (0.369) and well above ESM3, AminoAseed, and ProSST.
\protok{} and DPLM-2 both train their structure tokens jointly with sequence supervision while the other baselines do not, mirroring the inverse folding ordering.